\documentclass{article}

\usepackage{microtype}
\usepackage{graphicx}
\usepackage{subcaption}
\usepackage{booktabs} 
\usepackage{hyperref}

\usepackage{hyperref}

\newenvironment{Itemize}%
{\begin{itemize}%
\setlength{\itemsep}{0pt}%
\setlength{\topsep}{0pt}%
\setlength{\partopsep}{0pt}%
\setlength{\parskip}{0pt}}%
{\end{itemize}}
{\begin{enumerate}%
\setlength{\itemsep}{0pt}%
\setlength{\topsep}{0pt}%
\setlength{\partopsep}{0pt}%
\setlength{\parskip}{0pt}}%
{\end{enumerate}}
\usepackage{titlesec}
\titlespacing*{\paragraph}{0pt}{3.25ex plus 1ex minus .2ex}{1em}

\usepackage[preprint]{icml2026}

\usepackage{amsmath}
\usepackage{amssymb}
\usepackage{mathtools}
\usepackage{amsthm}

\usepackage{tabularx}
\newcommand{\grayrow}{\rowcolor[gray]{.90}}

\usepackage[capitalize,noabbrev]{cleveref}

\usepackage[table]{xcolor}
\definecolor{avgrow}{RGB}{220,245,220} 

\theoremstyle{plain}

\theoremstyle{definition}

\theoremstyle{remark}

\usepackage[textsize=tiny]{todonotes}

\icmltitlerunning{Physiology as Language: Translating Respiration to Sleep EEG}

\begin{document}

\twocolumn[
  \icmltitle{Physiology as Language: Translating Respiration to Sleep EEG}



  \icmlsetsymbol{equal}{*}

  \begin{icmlauthorlist}
    \icmlauthor{Kaiwen Zha}{equal,mit}
    \icmlauthor{Chao Li}{equal,mit}
    \icmlauthor{Hao He}{equal,mit}
    \icmlauthor{Peng Cao}{mit}
    \icmlauthor{Tianhong Li}{mit}
    \icmlauthor{Ali Mirzazadeh}{mit}
    \icmlauthor{Ellen Zhang}{mit}
    \icmlauthor{Jong Woo Lee}{bwh}
    \icmlauthor{Yoon Kim}{mit}
    \icmlauthor{Dina Katabi}{mit}
  \end{icmlauthorlist}

  \icmlaffiliation{mit}{MIT Computer Science \& Artificial Intelligence Laboratory}
  \icmlaffiliation{bwh}{Brigham and Women’s Hospital, Harvard Medical School}

  \icmlcorrespondingauthor{Kaiwen Zha}{kzha@mit.edu}

  \vskip 0.3in
]



\printAffiliationsAndNotice{\icmlEqualContribution}

\begin{abstract}

This paper introduces a novel cross-physiology translation task: synthesizing sleep electroencephalography (EEG) from respiration signals. To address the significant complexity gap between the two modalities, we propose a waveform-conditional generative framework that preserves fine-grained respiratory dynamics while constraining the EEG target space through discrete tokenization. Trained on over 28,000 individuals, our model achieves a 7\% Mean Absolute Error in EEG spectrogram reconstruction. Beyond reconstruction, the synthesized EEG supports downstream tasks with performance comparable to ground truth EEG on age estimation (MAE 5.0 vs. 5.1 years), sex detection (AUROC 0.81 vs. 0.82), and sleep staging (Accuracy 0.84 vs. 0.88), significantly outperforming baselines trained directly on breathing. Finally, we demonstrate that the framework generalizes to contactless sensing by synthesizing EEG from wireless radio-frequency reflections, highlighting the feasibility of remote, non-contact neurological assessment during sleep.

\end{abstract}
\section{Introduction}\label{sec:intro}

Physiological monitoring is central to modern healthcare, particularly for managing neurological and sleep disorders. The electroencephalogram (EEG), in particular, is the gold standard for measuring brain activity. It supports the diagnosis of conditions ranging from epilepsy and narcolepsy to PTSD and depression \cite{stephansen2018neural,klaming2022machine,acharya2015automated,newson2019eeg}, while also serving as a biomarker for drug toxicity and brain aging \cite{vancott2003drug,iosifescu2011electroencephalography,sun2019brain}.

However, the clinical utility of EEG is limited by the difficulty of acquisition. Standard polysomnography (PSG) relies on wet electrodes or tight headbands that are labor-intensive to apply and prone to artifacts. Worse, this cumbersome instrumentation can disrupt the very sleep it aims to measure \cite{agnew1966first}. Respiration, by contrast, is far easier to capture, requiring only non-invasive wearables such as breathing belts or even contactless radio-frequency (RF) sensors~\cite{yue2018extracting}. This gap—between the diagnostic value of EEG and the accessibility of breathing signals—motivates a fundamental question: \emph{Can we synthesize high-fidelity sleep EEG from breathing alone?}

Physiology traditionally treats the pulmonary and neurological systems as distinct domains. Yet growing evidence of respiratory--neurological coupling, particularly during sleep, suggests a deep, often hidden interdependence between them \cite{zelano2016respiration,heck2017breathing,kluger2021rest}. While these dependencies are complex and not well understood, we hypothesize that they encode shared latent information that modern machine learning can extract. If breathing carries a ``fingerprint'' of brain activity in the sleep state, then learning a mapping from respiration to EEG could democratize neurological monitoring, enabling scalable, comfortable, and potentially contactless assessment of brain activity.

We adopt the view that physiological time series, like language, can be described by a vocabulary of recurring patterns governed by biological ``grammar,'' and that different signals (e.g., EEG and breathing) may encode overlapping health information using distinct vocabularies. Based on this premise, we introduce a waveform-conditional generative model that translates breathing signals into sleep EEG. The model conditions on continuous respiratory waveforms, while treating EEG as a discrete language: it first learns an EEG vocabulary via tokenization, then trains a Transformer to translate respiration into EEG tokens using a masked-prediction objective. This asymmetric design preserves fine-grained continuous respiratory context while constraining the EEG search space, enabling high-fidelity generation.

We validate our approach at unprecedented scale using 14 sleep datasets spanning 28{,}394 individuals and 33{,}919 nights. Breathing-to-EEG translation achieves 7\% mean absolute error (MAE), preserving the key spectral and temporal structure of the ground-truth EEG. To assess utility beyond reconstruction, we evaluate the synthesized EEG on three downstream tasks---age regression, sex classification, and sleep staging---and benchmark against models trained on ground-truth EEG and on respiration alone. Across datasets, synthesized EEG approaches ground-truth performance (age MAE 5.0 vs.\ 5.1 years; sex AUROC 0.81 vs.\ 0.82; sleep-staging accuracy 0.84 vs.\ 0.88), while models trained directly on respiration perform substantially worse. This gap suggests that our model can disentangle and amplify brain-specific information latent in respiratory dynamics in a way that raw breathing signals do not.

Finally, we push toward non-invasive EEG monitoring using the MGH dataset, which includes synchronized RF-based breathing, belt-based breathing, and EEG. This unique setting allows us to demonstrate the feasibility of generating meaningful sleep EEG from wireless reflections alone: RF signals reflected off a sleeping body can be translated into EEG spectrograms at an accuracy comparable to those from a breathing belt (MAE of 8\% vs. 7\%), enabling neurological assessment without physical contact or wearables.

Our contributions are summarized as follows:

\vspace{-10pt}

\begin{Itemize}
    \item \textbf{A new translation task and a waveform-conditional framework:} We introduce cross-physiology translation from the pulmonary to the neurological domain during sleep and show that a waveform-conditional model with a tokenized EEG target enables accurate reconstruction and supports downstream tasks.
    \item \textbf{Feasibility of contactless EEG:} We provide the first demonstration of generating sleep EEG spectrogram from contactless wireless reflections, establishing a pathway toward remote, contact-free neurological assessment.
    \item \textbf{Large-scale clinical validation:} Across a diverse cohort spanning 12 datasets, we show that synthesized EEG is not only visually faithful but also computationally functional, achieving competitive performance on sleep staging, age estimation, and sex classification.
\end{Itemize}

\section{Related Works}

\textbf{Masked Generative Modeling.}
Our architecture builds on masked generative modeling, where representing data as discrete token sequences enables powerful, transferable models across domains. In NLP, masked reconstruction objectives (e.g., BERT \cite{devlin2019bert}, MASS \cite{song2019mass}, BART \cite{lewis2020bart}) support both representation learning and sequence-to-sequence generation. This paradigm has also been extended to vision, where tokenized images with masked prediction enable scalable generation, as shown by MaskGIT \cite{chang2022maskgit} and Fluid \cite{fan2024fluid}. We extend these ideas to waveform-conditional generation across physiological domains, translating continuous respiratory waveforms into tokenized EEG for cross-domain physiological synthesis.

\textbf{Cross-Modal Physiological Learning.} Prior multimodal learning on physiological data largely focuses on fusing signals for discriminative tasks~\cite{thapa_multimodal_2026,zhang_sensorlm_2025,abbaspourazad_large-scale_2024,deldari_crossl_2024,fang_promoting_2024}. More recently, researchers have explored generating one physiological modality from another to impute missing data or improve monitoring. However, existing work translates between modalities within the same physiological domain (e.g., neurological or cardiovascular), where signals are tightly coupled and share substantial information. In the neurological domain, NT-ViT \cite{lanzino2024neural} and diffusion-based methods \cite{eeg2fmri} translate EEG to fMRI to estimate spatially resolved hemodynamic activity from electrical recordings. In the cardiovascular domain, deep generative models translate PPG to ECG~\citep{sarkar2021cardiogan,lan2023performer,tang2023ppg2ecgps,ezzat2024ecg} and PPG to continuous arterial blood pressure (ABP) waveforms~\cite{ibtehaz2022ppg2abp}. In contrast, we are the first to demonstrate translation across physiological domains.

\textbf{Extracting Physiological Signals from Wireless Reflections.} There is growing interest in passive health monitoring, where a low-power radio device (akin to a WiFi router) emits signals and analyzes their reflections to infer physiological measurements without wearables~\cite{islam2022radar}. Prior work has used this paradigm to estimate a range of physiological and behavioral signals, including respiration~\cite{yue2018extracting,zhang2023pi}, heart rate~\cite{adib2015smart}, pose~\cite{zhao2018through}, stress~\cite{ha2021wistress}, activity \cite{liu2020human}, and even sleep stages \cite{he2025radio}. We extend this work by demonstrating, for the first time, that sleep EEG can be inferred from wireless reflections alone, enabling contact-free neurophysiological monitoring.

\begin{figure*}[ht]
\centering
    \includegraphics[width=\textwidth]{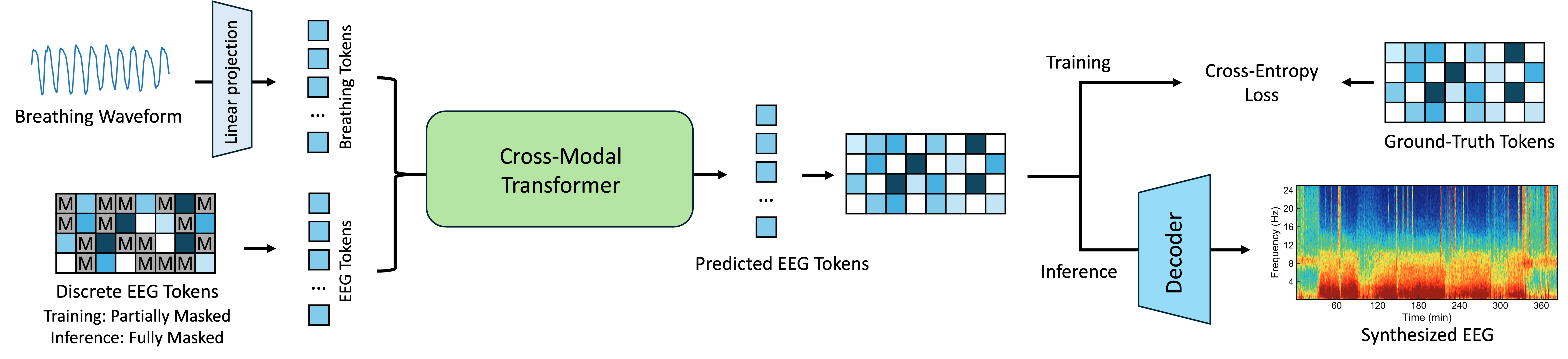}
    \captionof{figure}{\textbf{Model Pipeline.} 
    The model synthesizes sleep EEG from nocturnal breathing signals using an asymmetric embedding strategy to bridge the gap between modalities.
    The source breathing signal is processed as a raw waveform with a linear projection, while the target EEG is converted into discrete tokens by spectral transformation and vector quantization.
    A transformer-based model learns to translate the continuous respiratory context into the discrete neurological states using a masked generative modeling objective. 
    During inference, the model predicts the full sequence of EEG tokens from breathing alone, which are then decoded into an EEG spectrogram.
    }
    \label{fig:model-pipeline}

\end{figure*}

\begin{figure*}[ht]
\centering
    \includegraphics[width=\textwidth]{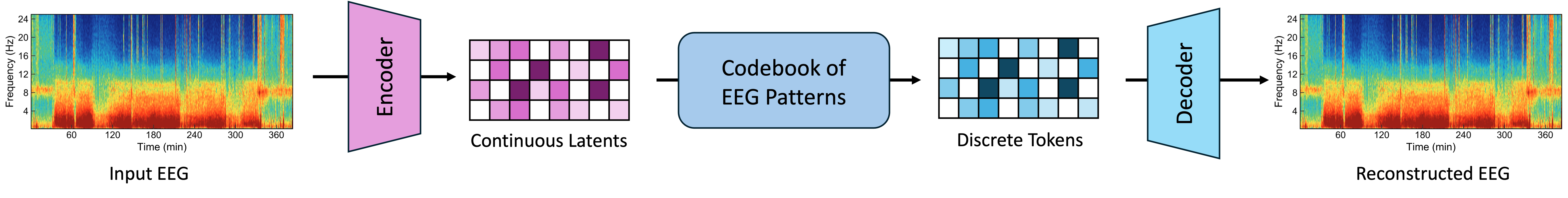}
\captionof{figure}{\textbf{Vector Quantization for EEG.} 
    The EEG spectrogram is discretized into tokens via a codebook of distinct EEG patterns. The resolution of the token ($4$ Hz $\times$ $4$ minutes per token) is chosen to align with the physiological semantics of sleep EEG.
    }
    \label{fig:tokenization}

\end{figure*}

\begin{figure}[ht]
    \centering
    \includegraphics[width=0.4\textwidth]{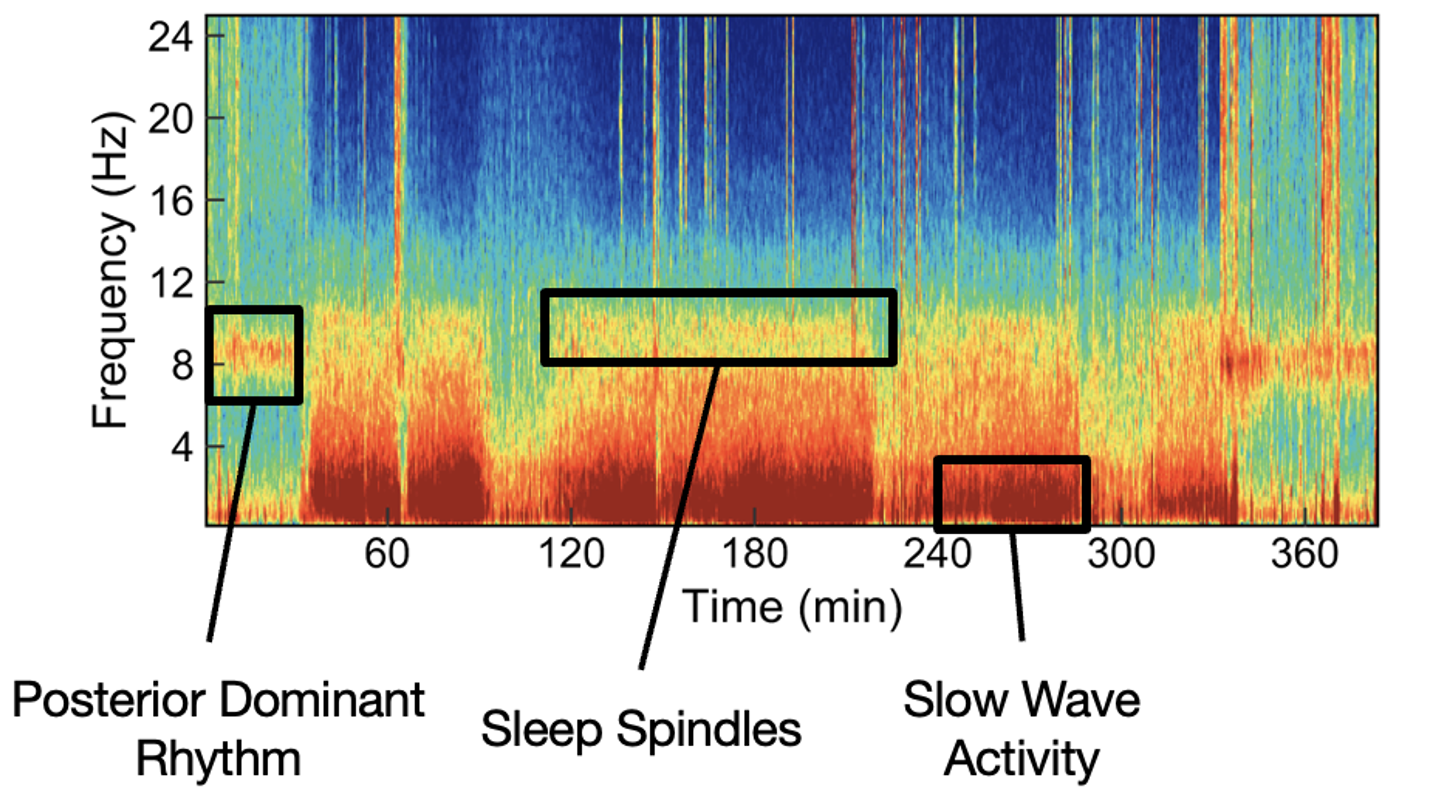}
    \captionof{figure}{\textbf{An example of sleep EEG spectrogram.}
    Key sleep patterns are annotated, including the posterior dominant rhythm (low-frequency activity around 8-12 Hz), sleep spindles (short bursts of 12-16 Hz activity), and slow-wave activity (prominent low-frequency oscillations below 4 Hz).
    }
    \label{fig:EEG}
\end{figure}

\vspace{-1mm}
\section{Method}
\label{sec:methodology}
We propose a cross-physiology translation framework to synthesize high-fidelity sleep EEG directly from nocturnal breathing signals. This task presents a fundamental challenge: the significant \textit{complexity gap} between the input and the target.
Respiratory signals are relatively simple, low-frequency waveforms driven by mechanical pulmonary effort, whereas EEG signals are complex, high-frequency, and stochastic representations of neurological activity.

We address this challenge by employing \textit{asymmetric} processing strategy for the two modalities to align their information density. For the source breathing signal, we prioritize the preservation of fine-grained context by encoding the raw waveform directly into a sequence of dense, continuous embeddings.
In contrast, for the target EEG, we constrain the high-dimensional search space by converting the signal into a time-frequency spectrogram and  discretizing it via a learned codebook of EEG patterns. This asymmetric embedding strategy enables us to leverage a transformer-based masked generative model to effectively translate the detailed, continuous respiratory context into the semantic, discrete vocabulary of brain activity. Our model pipeline is illustrated in Fig.~\ref{fig:model-pipeline}.

\subsection{Asymmetric Embedding}
\paragraph {Source: Raw Waveform Embedding.} 
We employ a minimal projection layer to preserve detailed physiological information in the raw waveform. Specifically, we divide the raw breathing signal $X_{\text{RESP}}$ into a sequence of non-overlapping 4-min segments and linearly map each segment into a continuous embedding space. This results in a sequence of continuous \textit{breathing tokens} that allows the subsequent Transformer to attend to fine-grained respiratory dynamics. This linear layer is trained together with the subsequent transformer.

This raw representation is critical. Our experiments revealed that utilizing heavy breathing encoders or time-frequency conversions significantly degraded performance, likely by filtering out subtle morphological cues or over-compressing the temporal context required to infer brain states.

\vspace{-10pt}

\paragraph{Target: Discrete Spectrogram Tokenization}
While the input must be preserved in raw form, the target EEG signal $X_{\text{EEG}}$ is noisy and its structure is distributed across frequencies, making direct waveform generation challenging. To make the search space tractable and capture the underlying EEG structure, we follow a two-step process:

\textit{1. Spectral Transformation}: We first convert the time-series single-channel EEG signal into a time-frequency spectrogram using the multitaper method~\cite{prerau2017sleep}. This approach provides Power Spectral Density (PSD) estimates with high spectral resolution and minimal leakage, capturing essential sleep EEG features such as posterior dominant rhythms ($8\text{--}12$ Hz), sleep spindles ($12\text{--}16$ Hz), and slow-wave activity ($0.5\text{--}1.5$ Hz) \cite{prerau2017sleep}, as shown in Fig.~\ref{fig:EEG}. We process the data using a window size of $30$ seconds; this duration is chosen to align with the standard epoch length defined by the AASM Manual for the Scoring of Sleep and Associated Events~\cite{berry2012aasm}.

 \textit{2. Vector Quantization (VQ)}: 
 We further compress the spectrogram into a sequence of discrete \textit{EEG tokens} using a VQGAN~\cite{esser2021taming}. This process effectively shrink the target search space, allowing the translation task to be a tractable classification problem over a finite vocabulary of physiological patterns instead of a complex regression.
 
As shown in Fig.~\ref{fig:tokenization}, the VQGAN is composed of a encoder $E$, a decoder $D$ and a learnable codebook $\mathcal{Z} = \{z_k\}_{k=1}^K \subset \mathbb{R}^{d}$. The encoder takes a spectrogram and produces a latent feature grid $\hat{z} = E(X_{\text{EEG}}) \in \mathbb{R}^{h \times w \times d}$, and then each feature vector is replaced by its nearest neighbor from the codebook,
$z_q = \text{arg min}_{z_k \in \mathcal{Z}} || \hat{z} - z_k ||_2$,
to obtain the quantized token sequence $z_q$. The decoder converts these tokens back to a spectrogram such that $\hat{X}_{\text{EEG}} = D(z_q) \approx X_{\text{EEG}}$.

We design the encoder's downsampling factor such that each token corresponds to a patch of approximately $4$ Hz in frequency and $4$ minutes in time. This resolution is strategically chosen to align the latent representation with the physiological semantics of sleep. The $4$ Hz frequency resolution naturally segments the spectral energy into canonical EEG bands, isolating Delta ($0\text{-}4$ Hz), Theta ($4\text{-}8$ Hz), Alpha ($8\text{-}12$ Hz), and Sigma ($12\text{-}16$ Hz) into distinct latent rows. The $4$-minute temporal resolution aggregates sufficient context to capture stable meso-scale sleep states.

The VQGAN is pretrained separately. In addition to the standard VQGAN objective~\cite{esser2021taming}, which jointly optimizes reconstruction fidelity, codebook commitment, and adversarial losses, we incorporate a correlation regularization term that explicitly maximizes the correlation between reconstructed spectrograms and the ground-truth targets. This encourages the tokenizer to preserve intrinsic physiological variability and spectral morphology, rather than converging to overly smooth or “average” patterns.

\subsection{Waveform-Conditional Generative Translation}
\label{method:masking}
With the asymmetric representations established—long-context raw embeddings for breathing and discrete semantic tokens for EEG—we formulate the synthesis as a sequence-to-sequence translation task. We employ a Transformer-based architecture and leverage a masked generative modeling approach~\cite{chang2022maskgit, li2023mage}.

During training, the transformer processes a concatenated sequence of breathing tokens and partially masked EEG tokens. We add learnable positional embeddings to preserve structure: 1D temporal embeddings for the breathing sequence and 2D embeddings (encoding time and frequency) for the EEG tokens. The model operates over this joint sequence to reconstruct the discrete tokens at masked positions, optimized via a cross-entropy loss. The masking ratio $\gamma$ is sampled from a truncated Gaussian distribution ($\mu=0.55$, clipped to $[0.5, 1.0]$) following \citet{li2023mage}. This variable masking schedule exposes the model to diverse difficulty levels, fostering robust contextual learning between the pulmonary and neurological modalities.

During inference, we synthesize EEG from a new breathing input by conditioning the model on breathing tokens concatenated with a \textit{fully masked} EEG sequence ($\gamma=1.0$). The predicted tokens are subsequently decoded by the frozen VQGAN to yield the final EEG spectrogram. Our experiments demonstrate that this masked generation strategy achieves superior reconstruction fidelity compared to sequential autoregressive generation.
\begin{table*}[t]
\setlength{\tabcolsep}{6pt}
\centering
\footnotesize
\renewcommand{\arraystretch}{1.2}

\caption{
\textbf{Mean Absolute Error (MAE) of EEG reconstruction from breathing signals across datasets and frequency bands.}
(a) Internal datasets are used during model development, while 
(b) External datasets are held out for evaluation only. 
Values are reported as mean ± standard deviation of MAE. 
\textbf{BB} denotes breathing from a wearable belts and \textbf{RF} denotes breathing from wireless reflections. 
Datasets marked with a dagger ($^{\dagger}$) include data from multiple sleep lab visits.}

\begin{subtable}[t]{\textwidth}
\subcaption{Internal datasets}
\centering
\begin{tabular}{lccccccc}
\toprule
\textbf{Dataset} & \textbf{Source} &
\textbf{Overall} &
$\boldsymbol{\delta}$ (0--4Hz) &
$\boldsymbol{\theta}$ (4--8Hz) &
$\boldsymbol{\alpha}$ (8--12Hz) &
$\boldsymbol{\sigma}$ (12--16Hz) &
$\boldsymbol{\beta}$ (12--32Hz) \\
\midrule
BWH & BB & 0.056 ± 0.017 & 0.062 ± 0.023 & 0.053 ± 0.022 & 0.054 ± 0.020 & 0.052 ± 0.018 & 0.056 ± 0.018 \\
SHHS$^{\dagger}$ & BB & 0.057 ± 0.020 & 0.057 ± 0.022 & 0.051 ± 0.022 & 0.058 ± 0.021 & 0.055 ± 0.024 & 0.058 ± 0.023 \\
MROS$^{\dagger}$ & BB & 0.066 ± 0.020 & 0.066 ± 0.025 & 0.062 ± 0.024 & 0.062 ± 0.022 & 0.060 ± 0.021 & 0.068 ± 0.020 \\
CHAT$^{\dagger}$ & BB & 0.067 ± 0.048 & 0.065 ± 0.045 & 0.069 ± 0.050 & 0.064 ± 0.052 & 0.066 ± 0.052 & 0.067 ± 0.050 \\
CCSHS & BB & 0.066 ± 0.015 & 0.072 ± 0.020 & 0.064 ± 0.021 & 0.064 ± 0.018 & 0.066 ± 0.016 & 0.067 ± 0.017 \\
NCHSDB & BB & 0.081 ± 0.036 & 0.080 ± 0.041 & 0.081 ± 0.040 & 0.076 ± 0.039 & 0.079 ± 0.038 & 0.083 ± 0.036 \\
P18C & BB & 0.104 ± 0.071 & 0.107 ± 0.072 & 0.102 ± 0.076 & 0.107 ± 0.080 & 0.102 ± 0.076 & 0.103 ± 0.070 \\
STAGES & BB & 0.089 ± 0.050 & 0.096 ± 0.054 & 0.086 ± 0.055 & 0.088 ± 0.055 & 0.087 ± 0.054 & 0.089 ± 0.051 \\
MGH & BB & 0.069 ± 0.026 & 0.074 ± 0.030 & 0.068 ± 0.032 & 0.070 ± 0.031 & 0.064 ± 0.029 & 0.067 ± 0.028 \\
MGH & RF & 0.076 ± 0.027 & 0.085 ± 0.038 & 0.075 ± 0.035 & 0.075 ± 0.031 & 0.070 ± 0.029 & 0.074 ± 0.027 \\
\midrule
\grayrow
\textbf{Average} &  & 0.068 ± 0.036 & 0.070 ± 0.039 & 0.065 ± 0.040 & 0.067 ± 0.039 & 0.065 ± 0.039 & 0.069 ± 0.037 \\
\bottomrule
\end{tabular}
\end{subtable}

\vspace{0.75em}

\begin{subtable}[t]{\textwidth}
\centering
\subcaption{External datasets}
\begin{tabular}{lccccccc}
\toprule
\textbf{Dataset} & \textbf{Source} &
\textbf{Overall} &
$\boldsymbol{\delta}$ (0--4Hz) &
$\boldsymbol{\theta}$ (4--8Hz) &
$\boldsymbol{\alpha}$ (8--12Hz) &
$\boldsymbol{\sigma}$ (12--16Hz) &
$\boldsymbol{\beta}$ (12--32Hz) \\
\midrule
WSC & BB & 0.059 ± 0.016 & 0.068 ± 0.022 & 0.053 ± 0.020 & 0.058 ± 0.018 & 0.055 ± 0.017 & 0.059 ± 0.017 \\
CFS & BB & 0.073 ± 0.019 & 0.089 ± 0.031 & 0.080 ± 0.030 & 0.070 ± 0.022 & 0.067 ± 0.019 & 0.070 ± 0.020 \\
MESA & BB & 0.073 ± 0.044 & 0.079 ± 0.053 & 0.072 ± 0.053 & 0.071 ± 0.050 & 0.069 ± 0.046 & 0.073 ± 0.043 \\
SOF & BB & 0.067 ± 0.020 & 0.075 ± 0.028 & 0.064 ± 0.026 & 0.063 ± 0.024 & 0.060 ± 0.022 & 0.067 ± 0.020 \\
UMASS & RF & 0.073 ± 0.024 & 0.092 ± 0.029 & 0.078 ± 0.029 & 0.080 ± 0.029 & 0.082 ± 0.032 & 0.066 ± 0.024 \\
\midrule
\grayrow
\textbf{Average} &  & 0.067 ± 0.030 & 0.075 ± 0.038 & 0.064 ± 0.038 & 0.064 ± 0.034 & 0.062 ± 0.032 & 0.066 ± 0.030 \\
\bottomrule
\end{tabular}
\end{subtable}
\label{tab:breathing_to_eeg_mae}
\end{table*}

\begin{table*}[t]
\centering
\footnotesize
\renewcommand{\arraystretch}{1.1}
\setlength{\tabcolsep}{6pt}
\caption{
\textbf{Signal-to-noise ratio (SNR) of EEG reconstruction from breathing signals across datasets and frequency bands.}
(a) Internal datasets used during model development and
(b) external datasets held out during training for evaluation.
Values are reported as mean ± standard deviation of SNR.
\textbf{BB} denotes breathing belts and \textbf{RF} denotes wireless breathing sensors.
Datasets marked with a dagger ($^{\dagger}$) include data from multiple sleep lab visits.}

\begin{subtable}[t]{\textwidth}
\centering
\caption{Internal datasets}
\begin{tabular}{lccccccc}
\toprule
\textbf{Dataset} & \textbf{Source} &
\textbf{Overall} &
$\boldsymbol{\delta}$ (0--4Hz) &
$\boldsymbol{\theta}$ (4--8Hz) &
$\boldsymbol{\alpha}$ (8--12Hz) &
$\boldsymbol{\sigma}$ (12--16Hz) &
$\boldsymbol{\beta}$ (12--32Hz) \\
\midrule
BWH & BB & 14.9 ± 1.9 & 17.0 ± 2.4 & 17.2 ± 2.8 & 16.6 ± 2.6 & 15.9 ± 2.4 & 13.6 ± 2.0 \\
SHHS$^{\dagger}$ & BB & 15.4 ± 1.8 & 19.0 ± 2.3 & 18.5 ± 2.6 & 17.0 ± 2.5 & 16.3 ± 2.4 & 13.8 ± 1.9 \\
MROS$^{\dagger}$ & BB & 14.1 ± 1.9 & 17.3 ± 2.4 & 16.3 ± 2.6 & 15.9 ± 2.6 & 15.0 ± 2.3 & 12.6 ± 2.0 \\
CHAT$^{\dagger}$ & BB & 14.5 ± 3.7 & 19.2 ± 3.7 & 17.0 ± 3.9 & 15.5 ± 4.1 & 14.0 ± 4.3 & 12.1 ± 4.3 \\
CCSHS & BB & 14.2 ± 1.9 & 17.7 ± 2.2 & 16.6 ± 2.7 & 15.3 ± 2.5 & 14.1 ± 2.1 & 12.3 ± 1.9 \\
NCHSDB & BB & 12.5 ± 2.6 & 16.7 ± 3.4 & 14.8 ± 3.2 & 13.4 ± 3.1 & 12.1 ± 2.9 & 10.5 ± 2.6 \\
P18C & BB & 11.7 ± 6.0 & 14.7 ± 5.5 & 14.0 ± 6.5 & 12.9 ± 6.7 & 12.3 ± 7.0 & 10.0 ± 6.8 \\
STAGES & BB & 11.7 ± 4.3 & 14.5 ± 4.2 & 13.9 ± 4.9 & 12.9 ± 4.9 & 12.1 ± 5.0 & 10.1 ± 4.7 \\
MGH & BB & 13.9 ± 2.2 & 16.7 ± 2.5 & 16.3 ± 3.0 & 15.3 ± 3.2 & 14.6 ± 2.8 & 12.2 ± 2.4 \\
MGH & RF & 12.6 ± 2.2 & 15.0 ± 2.9 & 14.7 ± 3.1 & 13.9 ± 3.0 & 13.4 ± 2.6 & 11.1 ± 2.2 \\
\midrule
\grayrow
\textbf{Average} &  & 14.1 ± 3.1 & 17.4 ± 3.4 & 16.6 ± 3.8 & 15.5 ± 3.7 & 14.6 ± 3.8 & 12.4 ± 3.4 \\
\bottomrule
\end{tabular}
\end{subtable}

\vspace{0.75em}

\begin{subtable}[t]{\textwidth}
\centering
\caption{External datasets}
\begin{tabular}{lccccccc}
\toprule
\textbf{Dataset} & \textbf{Source} &
\textbf{Overall} &
$\boldsymbol{\delta}$ (0--4Hz) &
$\boldsymbol{\theta}$ (4--8Hz) &
$\boldsymbol{\alpha}$ (8--12Hz) &
$\boldsymbol{\sigma}$ (12--16Hz) &
$\boldsymbol{\beta}$ (12--32Hz) \\
\midrule
WSC & BB & 14.3 ± 1.9 & 17.0 ± 2.1 & 17.6 ± 2.4 & 16.3 ± 2.4 & 15.5 ± 2.4 & 12.6 ± 2.2 \\
CFS & BB & 13.4 ± 2.1 & 15.7 ± 2.8 & 14.7 ± 3.0 & 14.7 ± 2.8 & 14.0 ± 2.5 & 12.2 ± 2.2 \\
MESA & BB & 13.5 ± 3.3 & 16.4 ± 3.6 & 15.6 ± 4.3 & 15.2 ± 4.3 & 14.3 ± 4.2 & 12.0 ± 3.7 \\
SOF & BB & 13.9 ± 2.2 & 16.4 ± 2.7 & 16.4 ± 3.1 & 16.0 ± 3.1 & 14.9 ± 2.9 & 12.4 ± 2.3 \\
UMASS & RF & 14.2 ± 1.7 & 16.6 ± 1.8 & 16.1 ± 2.2 & 14.7 ± 2.3 & 13.6 ± 2.2 & 12.4 ± 1.8 \\
\midrule
\grayrow
\textbf{Average} &  & 13.9 ± 2.5 & 16.6 ± 2.9 & 16.4 ± 3.5 & 15.7 ± 3.3 & 14.8 ± 3.3 & 12.3 ± 2.9 \\
\bottomrule
\end{tabular}
\end{subtable}

\label{tab:breathing_to_eeg_snr}
\end{table*}

\begin{figure*}[t]
    \centering
    \includegraphics[width=0.99\textwidth]{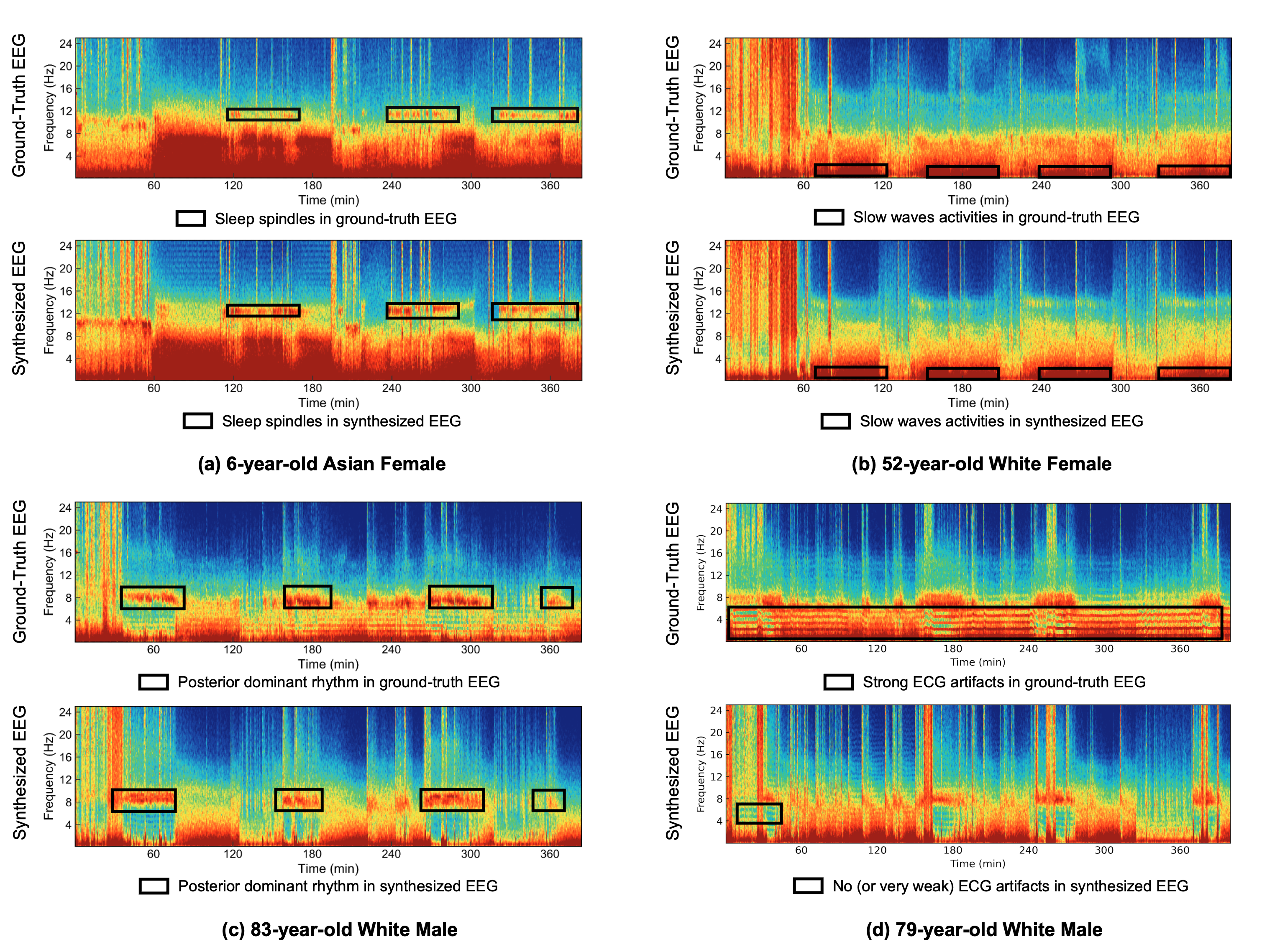}
    \vspace{-1mm}
    \captionof{figure}{\textbf{Visualization of EEG Reconstruction Results.}  In each panel, top row shows the ground-truth EEG, and bottom row shows the generated counterpart. The boxes highlight fine features in different EEG bands. These examples underscore the ability of the model to capture and replicate essential EEG features while eliminating some artifacts like the stripes in (d).}
    \vspace{-3mm}
    \label{fig:qualitative_results}
\end{figure*}

\section{Experiments}
\subsection{Evaluation Details and Downstream Tasks}
\paragraph{Reconstruction.} We evaluate EEG reconstruction by comparing the synthesized EEG spectrograms to their ground truth counterpart. We use two metrics: (1) \textbf{Mean absolute error (MAE)} computed on spectrogram values normalized to $[0,1]$ and (2) \textbf{Signal-to-noise ratio (SNR)}, measuring signal power relative to reconstruction error:
$\text{SNR} = 10 \log \frac{\lVert S \rVert^2}{\lVert S - \hat{S} \rVert^2}$.
We report both metrics for the full EEG spectrogram and for each canonical frequency bands: $\delta$ (0--4 Hz), $\theta$ (4--8 Hz), $\alpha$ (8--12 Hz), $\sigma$ (12--16 Hz), and $\beta$ (16--32 Hz).
Since no prior models reconstruct EEG from breathing, this task has no existing baselines for comparison. Thus, our evaluation focuses on reconstruction accuracy estimated via the above metrics. 

\vspace{-10pt}

\paragraph{Downstream Tasks.} We further evaluate the utility of the synthesized EEG spectrograms on three downstream tasks: age estimation, sex detection, and sleep-stage prediction. We compare models trained on synthesized EEG against two baselines: (1) \textbf{GT-EEG}: models trained on ground-truth EEG and (2) \textbf{Breathing}: models trained directly on raw breathing signals. This comparison quantifies the benefit of synthesizing EEG over using breathing alone, and how closely performance approaches that of ground-truth EEG.

All models take a full night of data as input. Age prediction is formulated as regression and trained with mean squared error (MSE) loss. Sex prediction is formulated as a binary classification and trained with binary cross-entropy loss. Sleep-stage prediction is posed as sequence labeling, predicting one label (Wake, Light, Deep, or REM) per 30-second segment and optimized with cross-entropy loss. 

Models operating on EEG use a ViT/Ti backbone ($\sim$5.7M parameters). For breathing baselines, we compare (1) a ResNet-style model~\citet{he2025radio} (current SOTA)  and (2) a hybrid transformer that prepends a convolutional encoder to the ViT/Ti backbone. The latter controls for architectural capacity, isolating gains from the synthesized EEG. 

For the prediction heads, we adapt the pooling strategy to the task level. For recording-level tasks (age and sex prediction), we apply global average pooling over the ViT token embeddings to obtain a single representation. Conversely, for sleep-stage prediction, we process the full sequence of embeddings to generate per-segment labels. Performance is reported using Mean Absolute Error (MAE) for age, AUROC for sex, and accuracy for sleep staging.

\vspace{-10pt}

\paragraph{Training and Cross Validation.}
We perform 4-fold, patient-wise cross-validation on the internal training datasets, assigning all nights from a given participant to the same fold. In each run, the model is trained on three folds and evaluated on the held-out fold, yielding subject-disjoint splits and preventing leakage from repeat visits. We use the same fold partitions for both VQGAN pretraining and translation model training.

\vspace{-10pt}

\paragraph{Datasets.}
We analyze a large-scale sleep corpus comprising 33{,}919 nights from 28{,}394 individuals (mean age 52.6 years; SD 25.1; range 3--102 years). Participants are 44.7\% female, and 72.8\% identify as White. The corpus aggregates 14 distinct sleep datasets: 9 are used for training and cross-validation, and 5 are held out as external test sets to evaluate out-of-distribution generalization. All 14 datasets include ground-truth EEG signals. Breathing signals are collected using a wearable belt in 12 datasets; one dataset (UMASS) uses RF-based breathing, and the MGH dataset includes paired RF-based and belt-based breathing. 

Different experiments focus on different datasets depending on availability of labels and input modality, or to match the datasets used to evaluate the baseline. Specifically:
\vspace{-2mm}
\begin{Itemize}
\item{\it Reconstruction from breathing belt} is evaluated on all datasets except UMASS, which lacks belt data. 
\item
{\it Reconstruction from RF} is evaluated on UMASS and MGH, the only datasets  containing RF data. 
\item
{\it Age estimation from breathing belt} is evaluated on CCSHS, CFS, CHAT, MESA, MGH, MrOS, SHHS, NCHSDB, SOF, and WSC, which have age labels.
\item
{\it Sex classification from breathing belt} is evaluated on CCSHS, CFS, MESA, MGH, SHHS, and WSC. Sex classification is evaluated for participants older than 12 years (post-puberty), excluding pediatric datasets, male- or female-only datasets, datasets with no sex labels, and individuals younger than 12 in the rest. 
\item  
{\it Sleep staging from breathing belt} is evaluated on MESA, MGH, SHHS-1, SHHS-2, and WSC.  This choice of datasets and the separation of SHHS into two datasets based on visit 1 vs. 2 allows for fair comparison with the SOTA in~\citet{he2025radio}.
\item 
{\it RF-based downstream tasks} are evaluated on MGH because it contains paired RF-based and belt-based breathing, allowing for comparing the two modalities.
\end{Itemize}
\vspace{-3mm}
Additional dataset details are in the Supplemental Material.

\subsection{Quantitative Reconstruction Results}
We evaluate the performance of our cross-physiology model for synthesizing sleep EEG from respiratory signal. 
Table~\ref{tab:breathing_to_eeg_mae} reports reconstruction performance in terms of MAE. On internal datasets used for training and cross-validation (Table~\ref{tab:breathing_to_eeg_mae}a), the model yields high reconstruction quality, achieving an average MAE of $0.068 \pm 0.036$ across EEG frequency bands. Errors are well balanced across bands, suggesting the model captures shared latent structure between respiratory dynamics and EEG activity without overfitting to a particular frequency range. 

Crucially, this performance transfers to unseen data. On the external datasets held out from training (Table~\ref{tab:breathing_to_eeg_mae}b), the model maintains comparable accuracy, with an average MAE of $0.067 \pm 0.030$. The close agreement between internal and external results demonstrates robustness to domain shift and supports the generalizability of the learned cross-modal mapping across datasets and sensing modalities.

Table~\ref{tab:breathing_to_eeg_snr} reports reconstruction performance in terms of SNR. Across datasets, the model achieves a strong average SNR of 14~dB, with values remaining consistent across cohorts and recording conditions. This stability mirrors the MAE results and further supports uniform reconstruction behavior and robust generalization to the external test datasets.

\subsection{Qualitative Reconstruction Results}
We qualitatively assess breathing-to-EEG synthesis by comparing generated EEG spectrograms to their paired ground-truth recordings. Fig.~\ref{fig:qualitative_results} shows representative examples. Across cases, the model preserves salient time-frequency structure and recovers canonical sleep-related patterns, including sleep spindles (Fig.~\ref{fig:qualitative_results}a), posterior dominant rhythms (Fig.~~\ref{fig:qualitative_results}c), and slow-wave activity (Figs.~\ref{fig:qualitative_results}a--d). We also observe that synthesized spectrograms are often visually cleaner than the recordings, with reduced artifacts. In particular, the model suppresses prominent ECG contamination visible as horizontal banding at harmonics of approximately 1~Hz in some recordings, which arises from electrode placement near vasculature (Fig.~\ref{fig:qualitative_results}d). 
Since this contamination is situational EEG sensor-level noise that is statistically independent of the respiratory signal, the model, conditioned purely on breathing dynamics, naturally excludes it, suggesting that the MAE and SNR reported above are conservative: some of the apparent “errors” may reflect benign differences introduced by denoising or artifact removal rather than true reconstruction errors.

\subsection{Impact of Demographics and Health Conditions}
\begin{figure}[t!]
    \centering
    \includegraphics[width=0.5\textwidth]{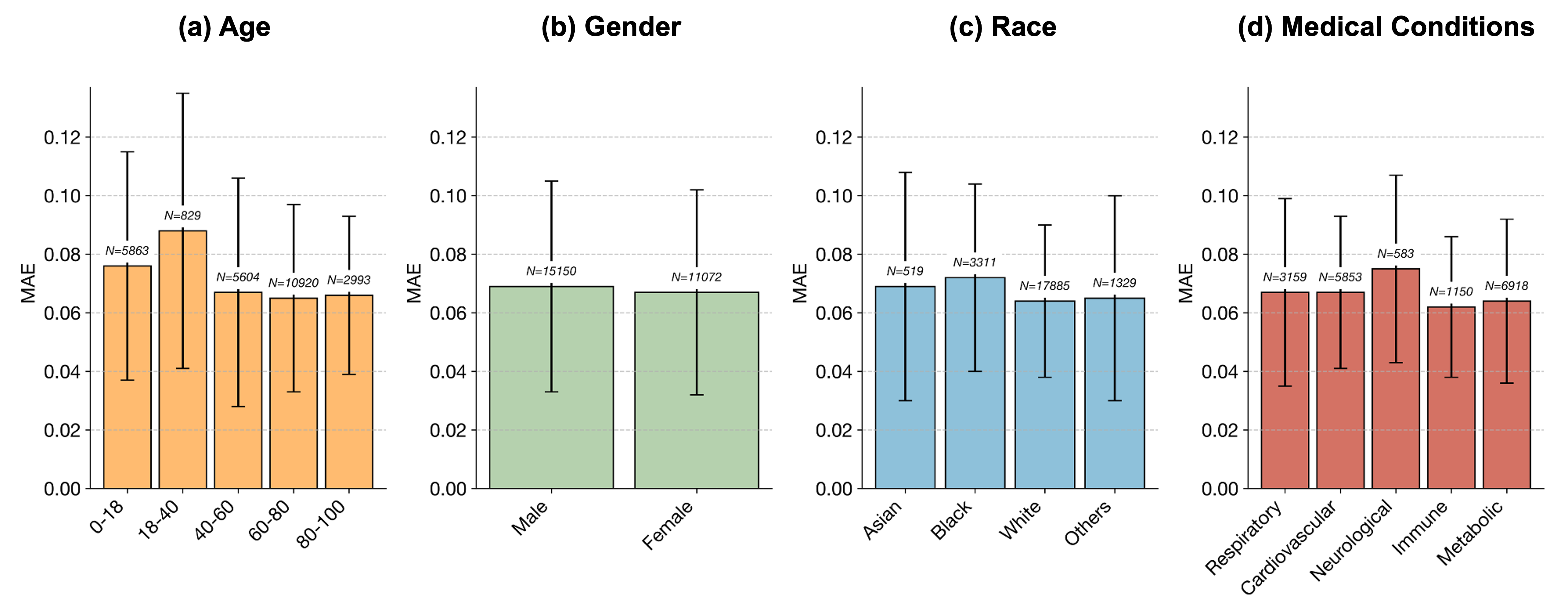}
    \vspace{-2mm}
    \captionof{figure}{\textbf{Performance across Demographics and Pre-existing Conditions.} Results are reported as Mean Absolute Error (MAE); lower is better. The figure shows that the MAE stays low across demographics and health conditions.} 
    \label{fig:dem_diseases}
    \vspace{-9mm}
\end{figure}

Figs.~\ref{fig:dem_diseases}a--c evaluate EEG synthesis accuracy (i.e., MAE) across subgroups defined by age (0--18, 18--40, 40--60, 60--80, 80--100 years), sex, and race (Asian, Black, White, Other). MAE is consistently low with only modest variation. Slightly higher MAE is observed in underrepresented sub-groups (ages 18--40; Black and Asian participants) and in children (0--18) who experience greater developmental EEG variability. We expect these gaps to diminish with larger training data from these sub-groups.

We also stratify by major disease categories (cardiovascular, autoimmune, metabolic, neurological, respiratory). As shown in Fig.~\ref{fig:dem_diseases}d, the MAE remains low across conditions with marginally lower error for cardiovascular and metabolic disorders and higher error for neurological conditions, likely reflecting smaller sample size.

\subsection{Performance on Downstream Tasks}

The utility of the synthesized EEG was further validated through age estimation, sex prediction, and sleep-staging. We compared these results against baselines trained on raw breathing signals and ground-truth EEG spectrograms.

\vspace{-10pt}

\paragraph{Age Estimation.}
Table~\ref{tab:age_prediction_by_dataset} shows age prediction from synthesized EEG closely matches ground truth (MAE 5.14 vs. 5.00 years), significantly outperforming direct prediction from breathing (MAE 7.66 years).

\begin{table}[t]
\centering
\caption{\textbf{Age prediction performance across datasets.}
Results are reported as Mean Absolute Error (MAE) in years ± Standard Deviation across the 4 folds; lower is better.}
\label{tab:age_prediction_by_dataset}
\resizebox{\columnwidth}{!}{
\scriptsize
\begin{tabular}{lcccc}
\toprule
\textbf{Dataset} & \multicolumn{2}{c}{\textbf{Breathing}} & \textbf{Synthesized EEG} & \textbf{GT-EEG} \\
 & CNN & ViT & ViT & ViT \\
\cmidrule(lr){2-3} \cmidrule(lr){4-4} \cmidrule(lr){5-5}
CCSHS   & 8.10 ± 0.55 & 6.11 ± 0.33 & 2.51 ± 0.65 & 3.97 ± 0.56 \\
CFS     & 13.3 ± 0.65 & 10.3 ± 0.55 & 7.34 ± 0.75 & 6.18 ± 0.34 \\
CHAT    & 4.07 ± 0.18 & 3.90 ± 0.24 & 2.17 ± 0.39 & 2.51 ± 0.19 \\
MESA    & 9.20 ± 0.16 & 8.14 ± 0.26 & 6.30 ± 0.26 & 6.87 ± 0.29 \\
MGH     & 12.2 ± 1.05 & 11.6 ± 0.56 & 7.96 ± 0.57 & 7.96 ± 0.67 \\
MROS    & 7.40 ± 0.12 & 4.86 ± 0.19 & 5.51 ± 0.72 & 5.07 ± 0.14 \\
NCHSDB  & 6.92 ± 0.63 & 6.19 ± 0.37 & 2.97 ± 0.40 & 3.02 ± 0.56 \\
SHHS    & 8.95 ± 0.29 & 8.44 ± 0.20 & 6.01 ± 0.06 & 6.21 ± 0.15 \\
SOF     & 9.31 ± 0.86 & 9.79 ± 1.18 & 3.85 ± 0.20 & 4.40 ± 0.91 \\
WSC     & 8.10 ± 0.43 & 7.28 ± 0.14 & 5.34 ± 0.16 & 5.16 ± 0.23 \\
\midrule
\textbf{Average}
 & \textbf{8.75 ± 0.492}
 & \textbf{7.66 ± 0.402}
 & \textbf{5.00 ± 0.416}
 & \textbf{5.14 ± 0.404} \\
\bottomrule
\end{tabular}
}
\end{table}

\vspace{-10pt}

\paragraph{Sex Prediction.}
 As shown in Table~\ref{tab:sex_by_dataset}, sex prediction from synthesized EEG nearly matches ground-truth EEG (AUROC=0.814 vs.\ AUROC=0.819), while breathing-based prediction is weaker (AUROC=0.768).
\begin{table}[t]
\centering
\caption{\textbf{Sex classification performance across datasets.} Results are reported as AUROC (male vs.\ female) ± Standard Deviation across the 4 folds; higher is better.}
\label{tab:sex_by_dataset}
\resizebox{\columnwidth}{!}{
\scriptsize
\begin{tabular}{lcccc}
\toprule
\textbf{Dataset} & \multicolumn{2}{c}{\textbf{Breathing}} & \textbf{Synthesized EEG} & \textbf{GT-EEG} \\
 & CNN & ViT & ViT & ViT \\
\cmidrule(lr){2-3} \cmidrule(lr){4-4} \cmidrule(lr){5-5}
CCSHS   & 0.706 ± 0.013 & 0.779 ± 0.040 & 0.773 ± 0.044 & 0.771 ± 0.044 \\
CFS     & 0.646 ± 0.079 & 0.768 ± 0.053 & 0.758 ± 0.043 & 0.820 ± 0.043 \\
MESA    & 0.749 ± 0.012 & 0.779 ± 0.026 & 0.866 ± 0.013 & 0.851 ± 0.013 \\
MGH     & 0.655 ± 0.032 & 0.705 ± 0.033 & 0.749 ± 0.038 & 0.780 ± 0.038 \\
SHHS    & 0.778 ± 0.024 & 0.801 ± 0.026 & 0.894 ± 0.025 & 0.864 ± 0.025 \\
WSC     & 0.768 ± 0.021 & 0.776 ± 0.010 & 0.845 ± 0.030 & 0.825 ± 0.030 \\
\midrule
\textbf{Average}
 & \textbf{0.717 ± 0.030}
 & \textbf{0.768 ± 0.031}
 & \textbf{0.814 ± 0.032}
 & \textbf{0.819 ± 0.032} \\
\bottomrule
\vspace{-2em}
\end{tabular}
}
\end{table}

\vspace{-10pt}

\paragraph{Sleep Stage Classification.}

\begin{table}[t]
\centering
\footnotesize
\caption{
\textbf{Sleep stage prediction performance across datasets.} 
Results are reported as night-level Accuracy ± Standard Deviation across the 4 folds; higher is better.
}
\label{tab:sleep_stages}
\setlength{\tabcolsep}{3pt}
\resizebox{0.86\columnwidth}{!}{
\begin{tabular}{lccc}
\toprule
\textbf{Dataset} & \textbf{\citet{he2025radio}} & \textbf{Synthesized EEG} & \textbf{GT-EEG} \\
\midrule
MESA   & 0.799 ± 0.101 & 0.851 ± 0.060 & 0.878 ± 0.077 \\
MGH    & 0.810 ± 0.080  & 0.827 ± 0.070 & 0.871 ± 0.063 \\
SHHS-1 & 0.788 ± 0.095  & 0.825 ± 0.069 & 0.877 ± 0.070 \\
SHHS-2 & 0.831 ± 0.063  & 0.841 ± 0.056 & 0.893 ± 0.053 \\
WSC    & 0.834 ± 0.084  & 0.850 ± 0.063 & 0.892 ± 0.048 \\
\midrule
Overall & 0.812 ± 0.085 & 0.839 ± 0.064 & 0.882 ± 0.062 \\
\bottomrule
\vspace{-2em}
\end{tabular}}
\end{table}

Sleep staging in clinical settings is typically performed from EEG, although recent work has shown that the hypnogram (a sequence of 30-second segments labeled Wake, Light, Deep, or REM) can be inferred directly from breathing. We compare the state-of-the-art breathing-based sleep staging model~\cite{he2025radio} to our two-stage approach: synthesize EEG from breathing and then predict sleep stages from the synthesized EEG. We also report an upper bound using ground-truth EEG. For a fair comparison, we evaluate on the datasets used in~\citet{he2025radio} and match model capacity across methods.

As shown in Table~\ref{tab:sleep_stages}, our breathing $\rightarrow$ synthesized EEG $\rightarrow$ staging pipeline improves over the breathing-based baseline (Acc.=0.839 vs. Acc.=0.812) and approaches the performance achieved using ground-truth EEG (Acc.=0.882). To our knowledge, this represents a new state of the art in sleep stage classification result from a breathing modality, narrowing the gap to EEG-based performance.
\vspace{-1mm}
\subsection{Results for EEG Synthesis from RF Signals}
We evaluate whether sleep EEG can be synthesized from contactless RF reflections acquired during sleep. We consider the RF-based cohorts in Tables~\ref{tab:breathing_to_eeg_mae} and~\ref{tab:breathing_to_eeg_snr} (MGH and UMASS). Across RF cohorts, synthesized EEG attains an average MAE of 0.075 and an average SNR of 13.4~dB, which is on par with belt-based breathing. We further test downstream utility on the MGH dataset, which includes paired belt- and RF-based breathing for the same people and nights, along with age, sex, and sleep stage labels.  As shown in Table~\ref{tab:wireless}, EEG synthesized from RF breathing supports age estimation, sex prediction, and sleep staging at slightly lower accuracy than EEG synthesized from belt breathing (age MAE 8.8 vs.\ 7.9 years; sex AUROC 0.70 vs.\ 0.74; sleep staging Acc. 0.81 vs.\ 0.82). Overall, these results show for the first time the feasibility of EEG inference from passive wireless sensing without wearable devices.

\vspace{-1.7mm}
\begin{table}[h]
\centering
\setlength{\tabcolsep}{6pt}
\footnotesize
\caption{\textbf{Performance of wireless signals compared to breathing belt (on MGH dataset) across three downstream tasks.}}
\label{tab:wireless}
\resizebox{0.8\columnwidth}{!}{
\begin{tabular}{lcc}
\toprule
Source & \textbf{BB} & \textbf{RF} \\
\midrule
Age (MAE) $\downarrow$      & 7.96 ± 0.57 & 8.83 ± 0.75 \\
Sex (AUROC) $\uparrow$  & 0.749 ± 0.038 & 0.703 ± 0.049          \\
Sleep Stage (Acc.) $\uparrow$ & 0.827 ± 0.070 & 0.814 ± 0.070          \\
\bottomrule
\end{tabular}}
\label{tab:wireless}
\end{table}

\vspace{-2mm}
\section{Conclusion}
We introduced the task of cross-physiology translation and showed that a waveform-conditional generative framework can decode the intricate, nonlinear coupling between the pulmonary and neurological systems to synthesize high-fidelity sleep EEG directly from breathing dynamics. Evaluated across 14 datasets (33{,}919 nights; ages 3--102), the method achieves low reconstruction error and generalizes across cohorts and sensing modalities. Notably, we provide the first evidence that contactless RF reflections can support the synthesis of meaningful EEG spectrograms, pointing toward remote, contact-free neurological assessment during sleep. Limitations include a remaining gap to ground-truth EEG and a focus on a single EEG channel (C4--A1), which can be addressed in future work. Overall, our results suggest a scalable path from ubiquitous breathing signals to EEG representations, enabling scalable remote neurophysiological monitoring without head-worn sensors.

\section{Impact Statement}
This work introduces a generative framework capable of translating respiratory dynamics into high-fidelity sleep EEG, utilizing a dataset of over 28,000 individuals. By demonstrating that detailed neurological information can be synthesized from breathing —including via contactless wireless reflections— this research holds the potential to democratize access to sleep EEG. It offers a scalable alternative to traditional polysomnography, potentially benefiting populations for whom wearable EEG is inaccessible or impractical, such as children, the elderly, and patients with sensory processing sensitivities.

{\bf Ethical Compliance and Data Governance.} All datasets utilized in this study were collected under strict ethical guidelines. Institutional Review Board (IRB) approval was obtained for all data collection protocols, and informed consent was secured from all participants prior to their inclusion in the source datasets. We strictly adhere to data use agreements that prohibit the re-identification of subjects.

{\bf Privacy and Dual-Use Concerns.} The capability to infer brain activity from contactless wireless signals raises distinct privacy considerations. Unlike wearables, radio-frequency (RF) sensing can theoretically be deployed unobtrusively. While our method requires a learned mapping that generalizes across populations, the possibility of inferring sensitive sleep architecture or biometric attributes (as demonstrated by our age and sex prediction results) from passive environmental signals underscores the need for robust privacy safeguards. Future deployment of such technology must be governed by frameworks that ensure transparency, explicit user consent, and privacy-by-design principles to prevent unauthorized surveillance.

{\bf Clinical Limitations and Safety.} While our generative model achieves high fidelity (7\% MAE) and strong downstream utility, synthesized EEG is a probabilistic reconstruction, not a direct measurement of neural voltage. Like all generative foundation models, there is a risk of hallucination, where the model might generate plausible but factually incorrect physiological features, or conversely, fail to capture rare pathological events not well-represented in the respiratory signal. Consequently, this technology should be viewed as a screening or support tool rather than a standalone diagnostic replacement. Clinical adoption will require rigorous human-in-the-loop validation to ensure that synthetic approximations are not mistaken for ground-truth physiology in critical care decisions.

\nocite{langley00}

\bibliography{example_paper}
\bibliographystyle{icml2026}

\clearpage
\appendix

\section{Supplementary Material}

\subsection{Dataset Summary}

\begin{table*}[h]
\centering
\small
\caption{\textbf{Summary of datasets used in this study.} External datasets (held out for testing only) are highlighted in grey. Dashes indicate unavailable data.}
\label{tab:dataset_summary}
\resizebox{\textwidth}{!}{
\begin{tabular}{lclcccccccc}
\toprule
\textbf{Cohort} &
\textbf{Usage} &
\textbf{Signal} &
\textbf{\#Part.} &
\textbf{\#Nights} &
\textbf{Female (\%)} &
\textbf{Age (mean $\pm$ SD)} &
\textbf{Asian} &
\textbf{Black} &
\textbf{White} &
\textbf{Other} \\
\midrule
BWH    & Internal & Belts    & 4866 & 4866 & 53.2 & 55.1 $\pm$ 17.1 & 3.5 & 14.1 & 65.7 & 16.7 \\
SHHS (v1 \& 2) & Internal & Belts & 5797 & 8444 & 52.4 & 64.5 $\pm$ 11.2 & 0.0 & 8.9  & 84.6 & 6.5 \\
MROS (v1 \& 2) & Internal & Belts & 2906 & 3930 & 0.0  & 77.6 $\pm$ 5.6  & 2.9 & 3.4  & 91.1 & 2.6 \\
CHAT   & Internal & Belts    & 1232 & 1639 & 52.2 & 7.0 $\pm$ 1.4   & 1.2 & 47.6 & 40.3 & 10.9 \\
CCSHS  & Internal & Belts    & 515  & 515  & 49.5 & 17.7 $\pm$ 0.4  & 0.2 & 35.9 & 59.6 & 4.3 \\
NCHSDB & Internal & Belts    & 3960 & 3960 & 43.1 & 8.8 $\pm$ 6.0   & 3.2 & 22.5 & 74.3 & 0.0 \\
P18C   & Internal & Belts    & 1983 & 1983 & 34.9 & 55.0 $\pm$ 14.3 & --  & --   & --   & --  \\
STAGES & Internal & Belts    & 1897 & 1897 & 52.2 & 45.9 $\pm$ 15.1 & 10.2& 4.2  & 78.4 & 7.2 \\
MGH    & Internal & Wireless & 881  & 881  & 43.2 & 54.4 $\pm$ 16.7 & 4.3 & 7.9  & 78.9 & 8.9 \\
\midrule
\grayrow
WSC    & External & Belts    & 1122 & 2569 & 45.9 & 59.8 $\pm$ 8.5  & 1.1 & 2.0  & 94.8 & 2.1 \\
\grayrow
CFS    & External & Belts    & 695  & 695  & 54.5 & 41.3 $\pm$ 19.4 & 0.0 & 55.4 & 41.4 & 3.2 \\
\grayrow
MESA   & External & Belts    & 2056 & 2056 & 53.6 & 69.4 $\pm$ 9.1  & 12.2& 27.8 & 36.1 & 23.9 \\
\grayrow
SOF    & External & Belts    & 453  & 453  & 100.0& 82.8 $\pm$ 3.1  & 0.0 & 8.4  & 91.6 & 0.0 \\
\grayrow
UMASS  & External & Wireless & 31   & 31   & --   & --              & --  & --   & --   & --  \\
\midrule
\textbf{Overall} & -- & -- & 28394 & 33919 & 44.7 & 52.6 $\pm$ 25.1 & 3.3 & 15.7 & 72.8 & 8.2 \\
\bottomrule
\end{tabular}
}
\end{table*}

We conducted a retrospective analysis using a dataset comprising 33,919 nights from 28,394 individuals (mean age 52.6 years, standard deviation 25.1, age range 3–102 years). Of these individuals, 44.7\% were female, and 72.8\% identified as white. The dataset was constructed by aggregating data from 14 distinct sleep datasets, of which 9 were used for training and cross-validation, while 5 datasets were reserved for external testing (refer to Table.~\ref{tab:dataset_summary}). 

The MGH dataset comprises polysomnography (PSG) recordings collected at the Massachusetts General Hospital (MGH) Sleep Laboratory between 2019 and 2022. Participants were required to be at least 18 years old, cognitively unimpaired, free from electronic implants, and not pregnant. Upon enrollment and screening, study coordinators installed wireless monitoring devices within the sleep laboratory and ensured connectivity to the clinic’s Wi-Fi network. A simplified floor plan of the monitored area was also created to support retrospective assessments of participant positioning. A total of seven sleep technicians independently annotated the dataset, each responsible for one PSG session. Overall, the dataset includes 881 PSG recordings from 881 distinct participants. All study procedures were approved by the Institutional Review Board (IRB) at the Mass General Brigham (IRM no. 2018P000337). The Massachusetts Institute of Technology (MIT) Institutional Review Board ceded review to the Mass General Brigham IRB. 

The UMass dataset comprises data collected during an observational home-based study conducted from 2019 to 2021, aimed at investigating wireless signal applications for monitoring movements and vital signs. Participants were required to be adults (18 years or older), have home Wi-Fi access, and be capable of providing informed consent or have consent provided by a legally authorized representative. They also agreed to confidential storage and use of the collected data, acknowledging that anonymized data might be utilized in scientific publications. The study protocol was reviewed and approved by the Massachusetts Institute of Technology Committee on the Use of Humans as Experimental Subjects (COUHES) (IRB no. 1910000024).

The other datasets were obtained from the publicly available  National Sleep Research Resource (NSRR) (\url{https://sleepdata.org/datasets}). 

The datasets encompassed two categories of breathing signal measurements. The first category includes nocturnal breathing data collected using wearable breathing belts (abdominal and thoracic) during polysomnography (PSG) sleep studies. The second category comprises nocturnal breathing signals collected via contactless sensing, utilizing a wireless radio-frequency sensor \cite{yue2018extracting} placed in sleep laboratories. This sensor captures breathing patterns by analyzing wireless signal reflections without physical contact or wearable devices. Combining these datasets enabled validation of model performance across different modes of breathing data acquisition and ensured robustness across varied study designs and demographic characteristics.

\subsection{Model and Training Details}
\subsubsection{Tokenization}
We use a CNN-based VQGAN encoder-decoder and quantizer to compress
$256\times512$ EEG spectrograms into discrete latent tokens and reconstruct them back. The
encoder consists of 5 blocks, each containing 2 residual layers. After each block, the feature
map is downsampled via average pooling with scale factors of $(4,2)$, $(2,2)$, $(2,2)$, and $(2,1)$.
The output is then quantized at each spatial location using a codebook with $8192$ entries,
each with a $32$-dimensional embedding. L2 normalization is applied to the latent codes
during quantization. The decoder mirrors the encoder structure, consisting of $5$ blocks with
$2$ residual layers each. After every block, the feature map is symmetrically upsampled to
progressively reconstruct the original resolution. The model is trained using the Adam optimizer with a learning rate of $4.8 \times 10^{-5}$ for $200$ epochs and a batch size of $120$. 

\subsubsection{Masked Generative Model}
For the masked generative model, we use ViT-Small as the transformer backbone. Both the encoder and decoder consist of $8$ transformer blocks with an embedding dimension of $768$ and $8$ self-attention heads.
The model is trained using cross-entropy loss on the masked EEG token positions for $500$ epochs using the AdamW optimizer with a peak learning rate of
$1.125 × 10^{-4}$, a batch size of $192$, and a weight decay of $0.05$. We apply linear warm-up learning rate scheduling over the first $40$ epochs, followed by cosine learning rate decay.

\subsubsection{Downstream Models}
The age and sex downstream models are trained for 33 epochs, while the sleep stage models are trained for 15 epochs. All models employ a 3-epoch learning rate warmup, with a batch learning rate of $1\times10^{-3}$ and a weight decay of $1\times10^{-2}$. For the contactless monitoring, a single epoch of fine-tuning is applied on wireless data with a learning rate of $5\times10^{-4}$ and a weight decay of $1$.

\subsection{Ablation of Generative Translation Approach}
In addition to the masked generative modeling approach, as described in Sec.~\ref{method:masking}, we also conducted ablation experiments using autoregressive (AR) generation to validate our choice of translation objective. We tested two  autoregressive ordering strategies: 

(1) \textbf{Sequential Autoregressive}: We flattened the 2D EEG token grid into a 1D sequence by ordering tokens from low to high frequency within each time step, and then advancing sequentially from left to right across time. This strictly causal formulation resulted in a 21.5\% increase in Mean Absolute Error (MAE) compared to the masked model.

(2) \textbf{Column-wise Parallel Autoregressive}: We grouped the vertical column of frequency tokens at each time step into a single unit and performed autoregressive prediction over these temporal chunks. This approach, while capturing instantaneous spectral structure better than the fully sequential baseline, still resulted in a 14.8\% increase in MAE.

We attribute the superiority of the masked approach to the inherent characteristics of the physiological translation task. 
Unlike natural language generation, synthesizing a sleep spectrogram from breathing is a \textit{global} translation problem, where the structural consistency across the entire night is essential. 
Autoregressive models are restricted to conditioning on past EEG tokens, making them vulnerable to error accumulation and unable to utilize future contexts. In contrast, the masked generative objective enables bidirectional attention, allowing the model to infer physiological states from both preceding and succeeding context to produce a holistically coherent spectrogram.

\subsection{More Examples of EEG reconstruction}
\begin{figure*}
    \centering
    \includegraphics[width=0.95\textwidth]{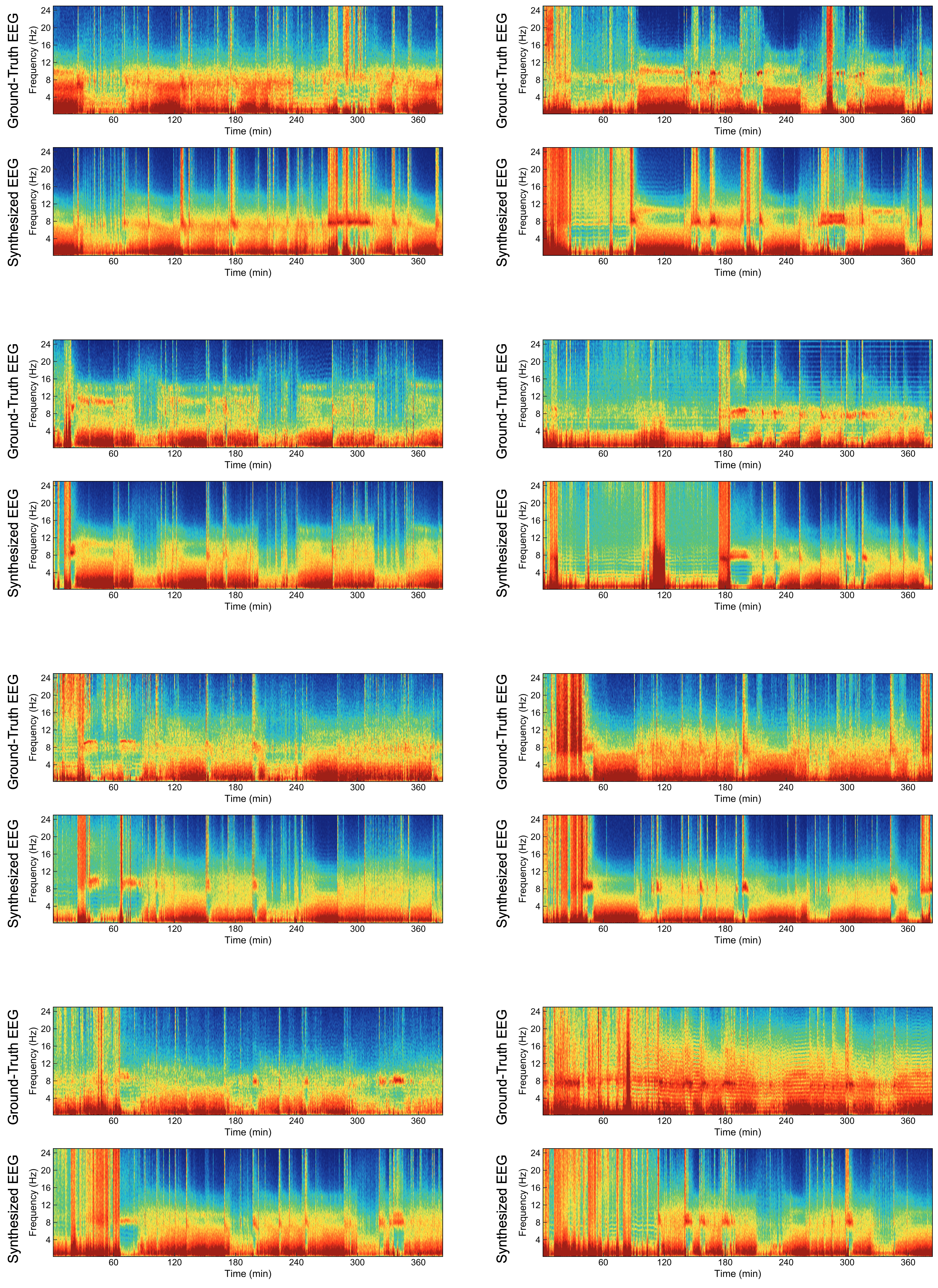}
    \caption{\textbf{More examples of EEG reconstruction.}}
    \label{fig:eg-1}
\end{figure*}

Additional examples of EEG reconstructions from breathing are visualized in Fig.~\ref{fig:eg-1}. In each group, the top row display the ground-truth EEG, while the bottom row show the generated counterpart.

\end{document}